\documentclass[letterpaper]{article} 
\usepackage{aaai25}  
\usepackage{times}  
\usepackage{helvet}  
\usepackage{courier}  
\usepackage[hyphens]{url}  
\usepackage{graphicx} 
\def\UrlFont{\rm}  
\usepackage{natbib}  
\usepackage{caption} 
\usepackage{algorithm}

\usepackage{algpseudocode} 
\usepackage{amsmath}

\usepackage{subcaption}
\usepackage{amsthm}
\usepackage{mathtools}

\newtheorem{theorem}{Theorem}
\newtheorem{definition}{Definition}

\usepackage{newfloat}
\usepackage{listings}
\DeclareCaptionStyle{ruled}{labelfont=normalfont,labelsep=colon,strut=off} 
\floatstyle{ruled}
\newfloat{listing}{tb}{lst}{}
\floatname{listing}{Listing}
\title{Iterative Counterfactual Data Augmentation}
\author{
    Mitchell Plyler\\
    Min Chi\\
}
\affiliations{
    Department of Computer Science, North Carolina State University\\
    mlplyler@ncsu.edu, mchi@ncsu.edu
    }

\begin{document}

\maketitle
\begin{abstract}
Counterfactual data augmentation (CDA) is a method for controlling information or biases in training datasets by generating a complementary dataset with typically opposing biases. Prior work often either relies on \emph{hand-crafted rules} or algorithmic CDA methods which can leave unwanted information in the augmented dataset. In this work, we show \textbf{iterative} CDA (ICDA) with initial, high-noise interventions can converge to a state with significantly lower noise. Our ICDA procedure produces a dataset where one target signal in the training dataset maintains high mutual information with a corresponding label and the information of spurious signals are reduced. We show training on the augmented datasets produces rationales on documents that better align with human annotation. Our experiments include six human produced datasets and two large-language model generated datasets.
\end{abstract}

%
\begin{links}
    \link{Code}{https://github.com/mlplyler/ICDA}
\end{links}

\section{Introduction}
Counterfactual data augmentation (CDA) is a method that can reduce targeted biases in training datasets, and ideally, reduces those biases in models trained on those datasets \cite{lu_gender_2020}. During CDA, counterfactuals are generated with roughly speaking the opposite bias of some original dataset. The original and counterfactual samples are concatenated into an augmented dataset where ideally their unwanted biases are balanced and canceled. In the literature, CDA can target specific biases through hand-crafted rules \cite{lu_gender_2020}, or general, unwanted biases with human annotators \cite{kaushik_learning_2020}. Both of these methods are costly either in expert knowledge or in annotator labor. Alternatively, cheaper, model-driven interventions can be noisy, both in where the interventions are made on source documents and how the interventions are made \cite{plyler_making_2021}. In this work, we leverage rationale networks \cite{lei_rationalizing_2016} to decide where to make counterfactual interventions \cite{plyler_making_2021}. 

Given a sample, rationale networks seek a \emph{subset} of the input with which to make a decision. Typically a portion of the network, the rationale selector, extracts some text, and another portion of the network, a classifier, makes a decision using the extracted text. The network learns to select subsets of text that make better predictions than other subsets of text \cite{lei_rationalizing_2016}.
Prior work \cite{chen_learning_2018} has shown that these networks are seeking the subset of text that maximizes the mutual information between the selected text and the prediction task. Often, this maximum mutual information (MMI) signal aligns with human reasoning and the ideal rationale network will find a policy that mimics the behaviour of human experts \cite{lei_rationalizing_2016}. With an aligned rationale model, we can use CDA to maintain a training dataset's information that aligns with human reasoning and we can reduce information that does not align \cite{plyler_making_2021}. This makes rationale networks a potentially ideal candidate for selecting where to make counterfactual interventions on documents.

Unfortunately, a core challenge for rationale networks are co-varying or spurious signals that cause the rationale network to converge to a policy where, in some cases, the spurious signals are used to make the prediction \cite{chang_game_2019}.
\citeauthor{plyler_making_2021} showed that CDA with rationale derived interventions can help lower the mutual information between spurious text and a target label, and help the rationale network get closer to the optimal, MMI policy. 
\citeauthor{plyler_making_2021} argued the benefits of CDA are dependent on the error rate of the rationale selector used to make the interventions, and empirically, they showed a second rationale model, trained on the augmented dataset, had a lower rationale error rate. If the second rationale model, the one trained on the augmented dataset, has a lower error rate than the initial model, we should see more benefits from CDA using that second rationale model instead of the initial. In fact, it stands to reason we can iteratively apply CDA with a new rationale model that is improving with each iteration.

This work builds on the ideas of rationalization \cite{lei_rationalizing_2016}, counterfactual data augmentation \cite{lu_gender_2020}, their combination \cite{plyler_making_2021}, and fixed-point processes to show the potential benefits of applying counterfactual data augmentation \textbf{iteratively}. Starting from an initial, noisy rationale model, we create a counterfactual dataset, train a new rationale network on the augmented data, and use that new network to create counterfactuals for the next iteration. We present information theoretic analysis, in a simplified setting, showing this iterative CDA (ICDA) algorithm forms a fixed-point process that should converge to a rationale network that better aligns with the maximum mutual information signal. Empirically, we show that our iterative process produces rationale models that align closely with human annotations. We perform experiments on six real, or human generated, datasets in the RateBeer and TripAdvisor settings. We also show that ICDA fits into the modern paradigm of generating a dataset using a large language model and training a light-weight classifier on that generated dataset. Across all eight experiments, ICDA outperforms the baselines.

\section{Method}
\subsection{Problem Definition}
Consider the typical supervised learning problem with inputs $X$ and output labels $Y$. We seek to train some model, $F$, that maps $Y \leftarrow F(X)$. A variation on the typical supervised problem is the rationale learning problem. A rationale model uses some subset of the input, $X_M \subseteq X$, to make prediction, $Y \leftarrow F(X_M)$. The rationale model is tasked with  
 \emph{learning which subset of input to use} to make the \textbf{same} prediction as using $X$. Typically, there is one signal in the input which would be ideal for the rationale model to select. Following \cite{plyler_making_2021}, we will call this subset $X_1$ and the label that corresponds with this subset $Y_1$. Often, $X_1$ is the subset of text that a human annotator would select as the \textit{rationale} for making the label prediction. Therefore, the quality of the rationale model can be assessed by measuring the agreement or alignment between the human annotated rationales $X_1$ and the selected rationales $X_M$. A successful rationale model therefore learns to select the subset $X_1$.
$X_M \leftarrow X_1$. \textbf{It is important to point out that this problem definition does not include the human annotations, ground-truth $X_1$, in the training dataset and they are only available in the test set for evaluation.} 

In this work, we assume there are multiple signals or aspects in the input and some of these signals are correlated with the target label $Y_1$. More specifically,   $X_1$ refers to the desired signals in the dataset  and $X_2$  represents another subset of input text that are  undesired or spurious signals. Often, we analyze the two aspect case where $X_1$ is desired and $X_2$ is considered spurious but correlated with the label. The general case would be for $N$ signals or subsets in the dataset which may or may not be disjoint. $X \leftarrow \{X_1, X_2, ..., X_N\}$. A typical multi-aspect example are hotel reviews were aspects within the reviews could refer to the hotel's location, cleanliness, service, etc.

\subsection{Background: Noisy CDA} \label{sec:beta}
 \citeauthor{chen_learning_2018} showed these rationale networks are seeking the subset or signal in the training dataset that maximizes the mutual information with the labels under some constraints $G$. These constraints are typically over the size or contiguity of the rationales. This is referred to as the maximum mutual information (MMI) criteria \cite{chen_learning_2018}.

\begin{equation}
\label{eq:mmi}
\max_{G}I(X_M;Y) \;\;\; \mbox{subject to} \;\;\; M \sim G(X)     
\end{equation}

In our problem definition, the rationale network will ideally learn $X_M \leftarrow X_1$. For the typical rationale network to be applicable, we are assuming that $X_1$ is the most informative signal in the input. We also assume there is high mutual information between the spurious signal and the target label, $I(X_2,Y_1)$. Typically, the dimensionality of the problem precludes an exhaustive search over subsets, so the rationale network seeks this MMI solution through Monte Carlo \cite{lei_rationalizing_2016}. The Monte Carlo approximation, and the high correlation or mutual information between $X_2$ and $Y_1$ means that our rationale network sometimes makes the mistake of selecting $X_2$ instead of $X_1$. The central hypothesis of \cite{plyler_making_2021} is that reducing the mutual information between $X_2$ and $Y_1$, $I(X_2,Y_1)$ will help the rationale network in identifying the correct relationship $X_M \leftarrow X_1$. 
\citeauthor{plyler_making_2021} showed that with perfect CDA, the spurious mutual information, $I(X_2,Y_1)$ would be eliminated in the augmented dataset. Perfect CDA would require perfect knowledge of $X_1$ and would make the process self-redundant. They reasoned that lowering mutual information between the spurious signal and the label more than the lower the mutual information between the target label and the target text would help the rationale network. They first defined the change in mutual information $\Delta I^a_{X,Y}$  from the original dataset  (X, Y) to the augmented dataset ($X^a$, $Y^a$). 
\begin{equation}
 \Delta I^a_{X_i,Y_j} = I(X_i,Y_j) - I(X_i^a,Y_j^a)    
\end{equation}
They then defined conditions when CDA will be successful:
\begin{equation}
 \label{eq:Idelta}
     \Delta I^a_{X_2,Y_1} - \Delta I^a_{X_1,Y_1} >0
\end{equation}
We adopt the same definition of success in this work. In their work, they proposed a three stage approach: train an initial rationale selector, generate a counterfactual dataset, and finally train a second rationale selector on the augmented dataset. They reasoned that if Equation \ref{eq:Idelta} was satisfied, the second rationale selector should be better at identifying the target signal than the first. They analyzed an error model where the first or initial selector made the mistake of selecting and modifying the spurious signal $X_2$ instead of the original signal $X_1$ at an error rate $\alpha$. In the augmented dataset, they derived conditional probabilities dependent on the initial conditional probabilities in the original dataset and the error rate of the initial selector, $\alpha$.

\citeauthor{plyler_making_2021} argued that the benefits of CDA are dependent on the error rate of the rationale selector and showed that there is a minimum error rate, $\alpha$, necessary for CDA to be beneficial. When the target signal is more informative than the spurious signal, $I(X_1,Y_1)>I(X_2,Y_1)$, the minimum error to see CDA benefits is actually pretty small. In fact, the initial selectors from \citeauthor{plyler_making_2021} imply CDA should not be beneficial.

\subsection{Helpful Counterfactual Generation Errors}
In this section, we will demonstrate how a suboptimal counterfactual generation process can, in fact, be advantageous for CDA by increasing the minimum error budget, thereby enhancing the benefits derived from CDA.
\begin{figure}
\includegraphics[width=0.45\textwidth]{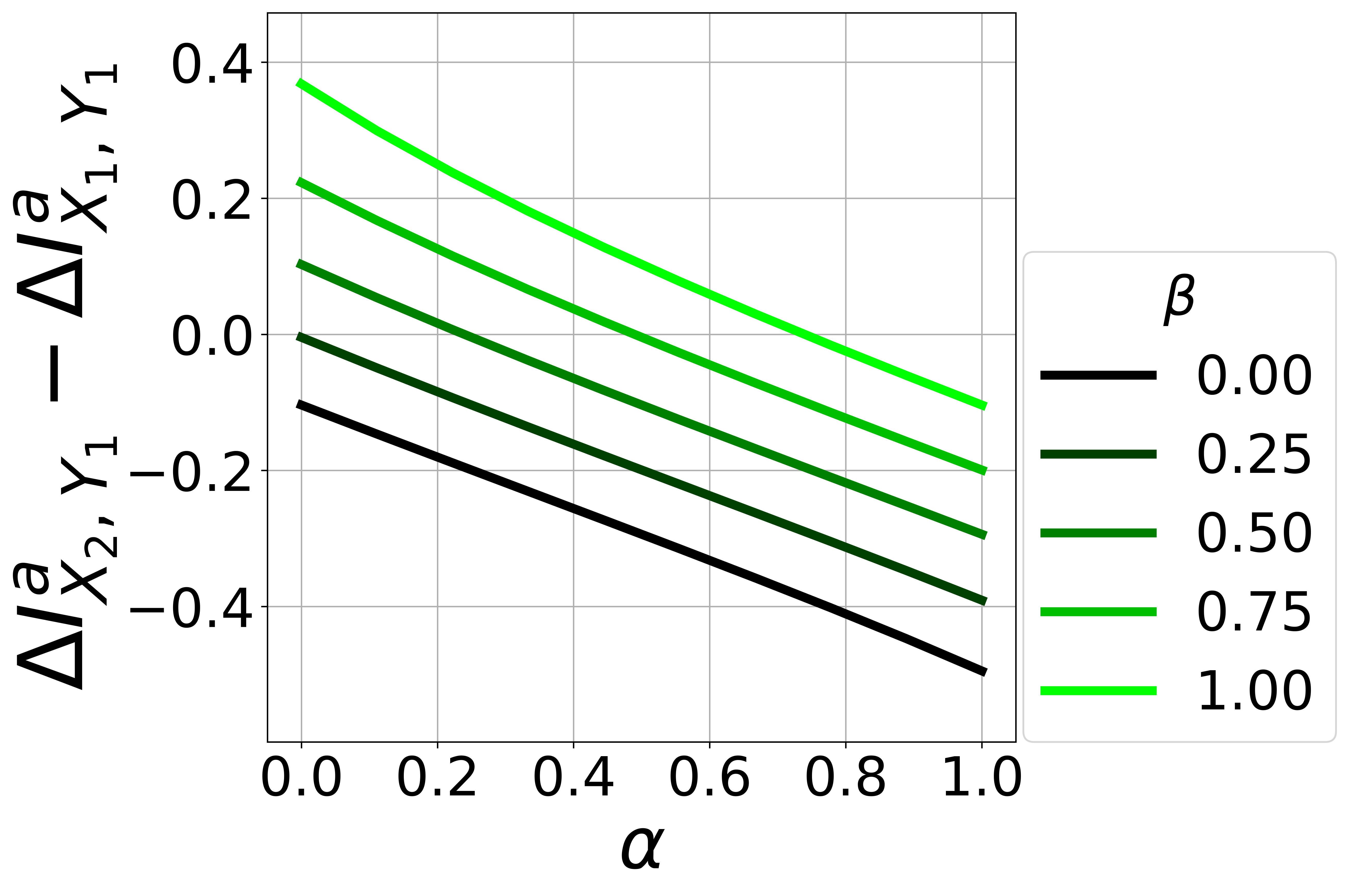}
\centering
\caption{Increasing $\beta$ helps CDA.}
\label{img:dI_beta}
\end{figure}
\citeauthor{plyler_making_2021}'s error or noise analysis was of course an approximation and it focused solely on the worst-case scenario: CDA perfectly flips $X_2$ when generating the counterfactual based on an incorrect rationale. In this section, we step back from the worst-case scenario by first considering the situation where $X_2$ is not perfectly flipped when it is modified during counterfactual creation.

Consider the example document 
\[ \textit{This beer smells great. It tastes fantastic.}\]
If we consider $Y_1$ our desired label corresponding to the smell aspect, then the subset of text that we should consider for this document would be $x_1=\textit{This beer smells great.}$ and the subset that we label spurious $x_2 = \textit{It tastes fantastic.}$ Consider the $\alpha$ noise from \citeauthor{plyler_making_2021}, we incorrectly select the spurious subset $X_2$ and we change its content to match the flipped label. This error would produce the counterfactual document:
\[\textit{This beer smells great. It tastes terrible.} \]
Now, consider the scenario where we select the incorrect portion of text $x_2$ but we still flip its sentiment correctly according to $X_1$ to match the flipped label.
\[ \textit{It tastes fantastic} \rightarrow \textit{It smells terrible.}\]
The new counterfactual document would read:
\[\textit{This beer smells great. It smells terrible.} \]
This document is obviously nonsensical, and it adds noise to our augmented dataset, but we will show this error is not as harmful as previous analysis where CDA flipped the taste sentiment.

Our augmented dataset now has two documents:
\[<positive> \textit{This beer smells great. It tastes fantastic.}\]
\[<negative> \textit{This beer smells great. It smells terrible.}\]
In half the augmented examples, we have the original conditional distributions: $P(Y_1|X_1)$ and $P(Y_1|X_2)$ just from including the original, unaltered sample. In the other half of the augmented examples, we have: $P(Y_1|X_1)=P(Y_1)$ and $P(Y_1|X_2)=P(Y_1)$ in the counterfactual samples. 
Lets say this happens only when we select the incorrect text which happens with probability $\alpha$, and when we flip the sentiment of the selected text correctly which we say happens with probability $\beta$.

When \citeauthor{plyler_making_2021} defined conditional probabilities in the augmented dataset, $P(Y_1^a|X_1^a)$ and $P(Y_1^a|X_2^a$,
they had an $\alpha$ portion of the documents that defined the error case and a $(1-\alpha)$ portion of the documents that defined the non-error case. Here we refine the analysis and define the error case for $X_1$ as 

\begin{multline}
P(Y_1^a|X_1^a)_{\text{error}} = \alpha \Bigl( (1-\beta)P(Y_1)  \\
+ \beta \Bigl( \frac{1}{2}P(Y_1) + \frac{1}{2} P(Y_1|X_1) \Bigr) \Bigr)
\end{multline}

This shows for the $\alpha$ portion of documents, where we grab the correct text, with probability $(1-\beta)$, we change the taste text as before and we have $P(Y_1)$. With probability $\beta$, we change the taste text to have negative smell sentiment, so half of the augmented dataset has no mutual information with the label $P(Y_1)$ and half maintain the information with the label $P(Y_1|X_1)$. Of course, our analysis should acknowledge the possibility that the rationale is correct, $X_1$ is modified, but the counterfactual is in correct, $1-X_2$ is inserted. Therefore, we will say that the correct $1-X_1$ is inserted with probability $\beta$ and the incorrect text is inserted with probability $1-\beta$. We assume that the inserted $X$ matches the counterfactual label $1-Y_1$. We can add back in the correct portion of the augmented dataset, $(1-\alpha)P(Y_1|X_1)$ and simplify to find the relation. We can repeat the analysis for $P(Y_1|X_2)$ to find the augmented conditional probabilities:

\begin{equation}
\label{eq:new_augmented_cond}
\begin{split}
  P(Y_1^a|X_1^a) &= \left( \frac{\alpha}{2} + \frac{1}{2} - \frac{\beta}{2} \right) P(Y_1)  \\
  &\quad + \left( -\frac{\alpha}{2} + \frac{1}{2} + \frac{\beta}{2} \right) P(Y_1|X_1),  \\
  P(Y_1^a|X_2^a) &= \left( -\frac{\alpha}{2} + \frac{1}{2} + \frac{\beta}{2} \right) P(Y_1) \\
  &\quad + \left( \frac{\alpha}{2} + \frac{1}{2} - \frac{\beta}{2} \right) P(Y_1|X_2).
\end{split}    
\end{equation}

To analyze the problem, we approximate using binary variables \cite{plyler_making_2021}: $p(Y_1|X_1)=.95$, $p(X_1)=p(X_2)=p(Y_1)=\frac{1}{2}$. Figure \ref{img:dI_beta} shows this $\beta$ error affects the success criteria of CDA. Remember the intersection of the x-axis shows the break even point of CDA and everything on the positive y-axis is a benefit while everything below the x-axis is a detriment. As our $\beta$ error increases, we are actually increasing our error budget for seeing benefits from CDA. This suggests a degenerate counterfactual generator that only injects $X_1|1-Y_1$ regardless of the context of the document, $X_2$, can actually be beneficial from an information theoretic view. We use this insight in section \ref{sec:beta} to strategically select counterfactual examples that increase our $\beta$ rate.

\subsection{Iterative Counterfactual Data Augmentation} 
\label{sec:iter}

\begin{algorithm}[h!]
\caption{Iterative CDA Procedure}\label{alg:iter_cda}
\begin{algorithmic}
\Require $D$ is a dataset with documents $X$ and labels $Y$.
\State $D' \gets D$
\While {not converged}
    \State $S \gets train\_selector(D')$ 
    \State $D^c \gets infer\_counterfactuals(D,S)$ 
    \State $D^a \gets concatenate(D,D^c)$
    \State $D' \gets D^a$
\EndWhile
\end{algorithmic}
\end{algorithm}

\citeauthor{plyler_making_2021} argued the benefits of CDA are proportional to the error rate, $\alpha$, of the initial rationale selector $S^{k=0}$. If training on counterfactually augmented data yields a lower error rationale selector, why not use that new rationale selector for another round of CDA? This question motivates our iterative approach where with each CDA iteration we train a better rationale selector and we lower the error rate of our counterfactual interventions.

Algorithm \ref{alg:iter_cda} outlines the ICDA method. We initialize our training dataset $D'$ with some unaugmented, source dataset $D$. We train a rationale selector $S^k$ on $D'$ using both the original dataset $D$ and the selector $S^k$ to infer a set of counterfactuals $D^c$ for the $k$th iteration. The augmented dataset $D^a$ is the concatenation of the original dataset $D$ and the counterfactual dataset $D^c$. The augmented dataset $D^a$ becomes our training dataset $D'$ for the next iteration. In this section, we will show that the error rate $\alpha$ of the selector $S$ decreases with each iteration.

In our ICDA procedure,  the error rate of the $k+1$th iteration's selector $S^{k+1}$, $\alpha^{k+1}$, is dependent on the error rate of $S^{k}$, $\alpha^{k}$:
\begin{equation}\label{eq:iter_alpha}
\alpha^{k+1} = \psi(\alpha^{k}) 
\end{equation}
Where $\psi$ is our iterative operator and consists of the functions $infer\_counterfactuals(D,S^k)$ and $train\_selector(D')$ from Algorithm \ref{alg:iter_cda}. To illustrate the convergence of process\ref{eq:iter_alpha} and Algorithm \ref{alg:iter_cda}, we will revisit the simplified scenario involving binary random variables. In this binary variable context, we will define our rationale scheme as follows:

\begin{definition} \label{def:binary_rat}
    Given random variables $X_1$, $X_2$, and $Y_1$ along with $n$ observations of these variables, our MMI rationalization scheme selects the variable in  $X$ that maximizes the agreement with expected error $\alpha$.
\end{definition}

First notice the error rate of the rationale selector in scheme \ref{def:binary_rat} is dependent on the difference in mutual information of the spurious signal and the target signal, $\delta$. We will call this relation operator $R$:
\begin{equation}
\delta \coloneqq I(X_1,Y_1) - I(X_2,Y_1)
\end{equation}
\begin{equation}\label{eq:R}
    \alpha^{k+1}=R(\delta,n)
\end{equation}

An increase in $\delta$ would increase the gap between $P(Y_1|X_1)$ and $P(Y_1|X2)$ and therefore decreases the rate of incorrectly choosing $X_2$ which is the expected error $\alpha^{k+1}$. Also note the strength of the rationale selector is dependent on the number of observations $n$. With enough observations, the rationale selector correctly identify the subset of $X$ with the highest mutual information. For any positive $\delta$, $\alpha^{k+1} \rightarrow 0$ as $n \rightarrow \infty$.

\begin{figure}
     \centering
     \begin{subfigure}[b]{0.2\textwidth}
         \centering
         \includegraphics[width=\textwidth]{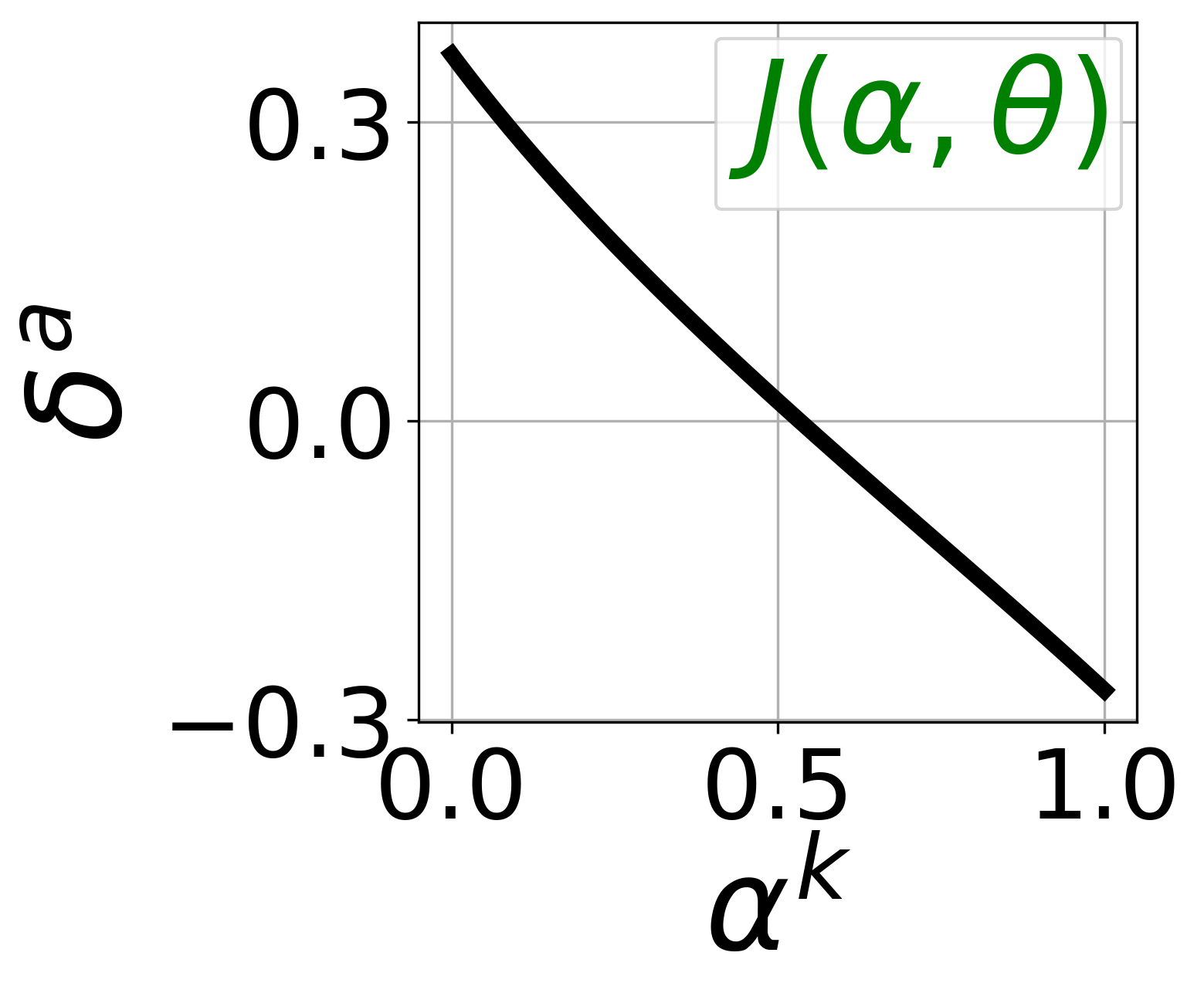}
         \caption{}
     \end{subfigure}
     \begin{subfigure}[b]{0.2\textwidth}
         \centering
         \includegraphics[width=\textwidth]{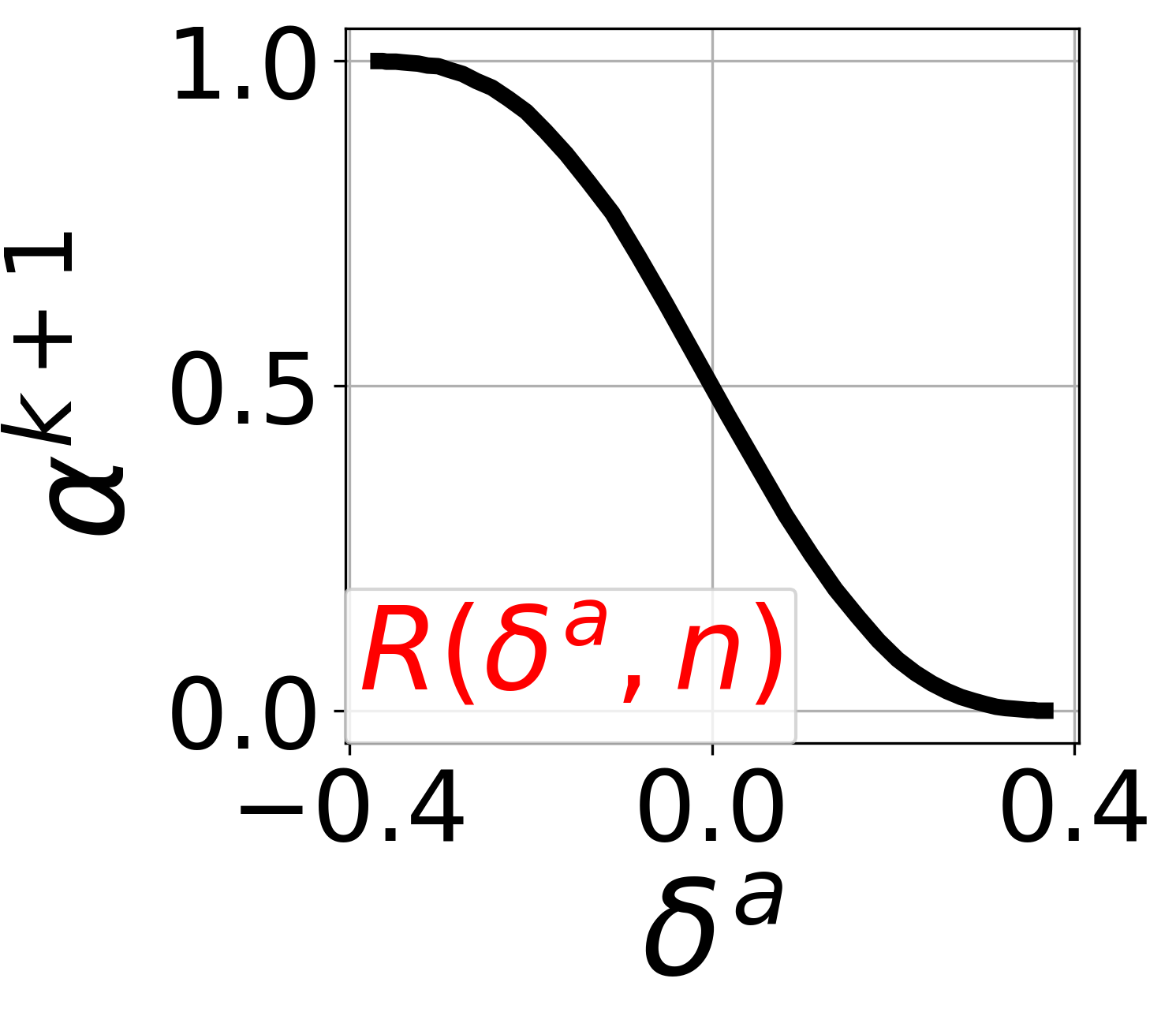}
         \caption{}
     \end{subfigure}     
     \begin{subfigure}[b]{0.2\textwidth}
         \centering
         \includegraphics[width=\textwidth]{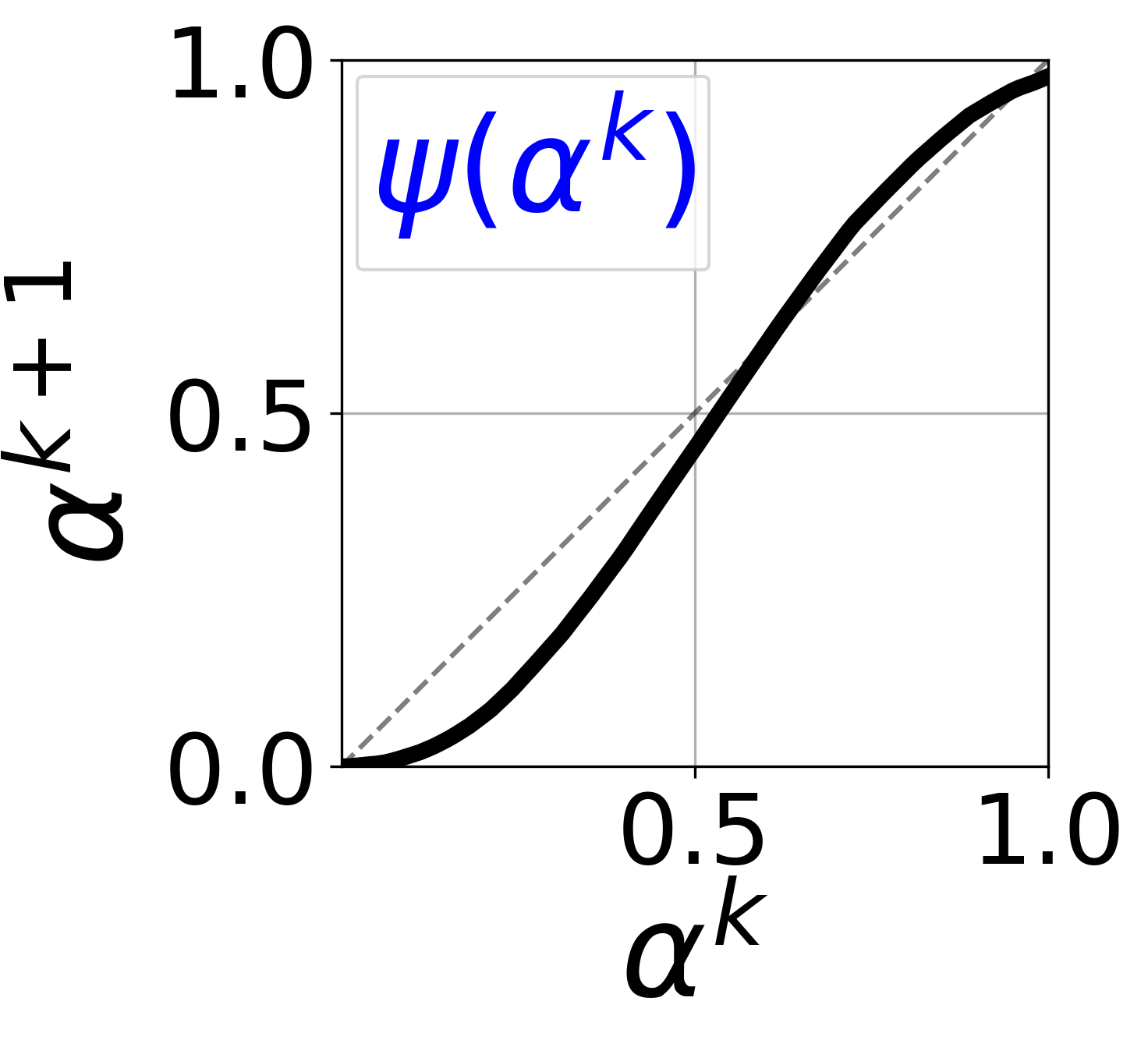}
         \caption{}
    \end{subfigure}               
    \begin{subfigure}[b]{0.2\textwidth}
         \centering
         \includegraphics[width=\textwidth]{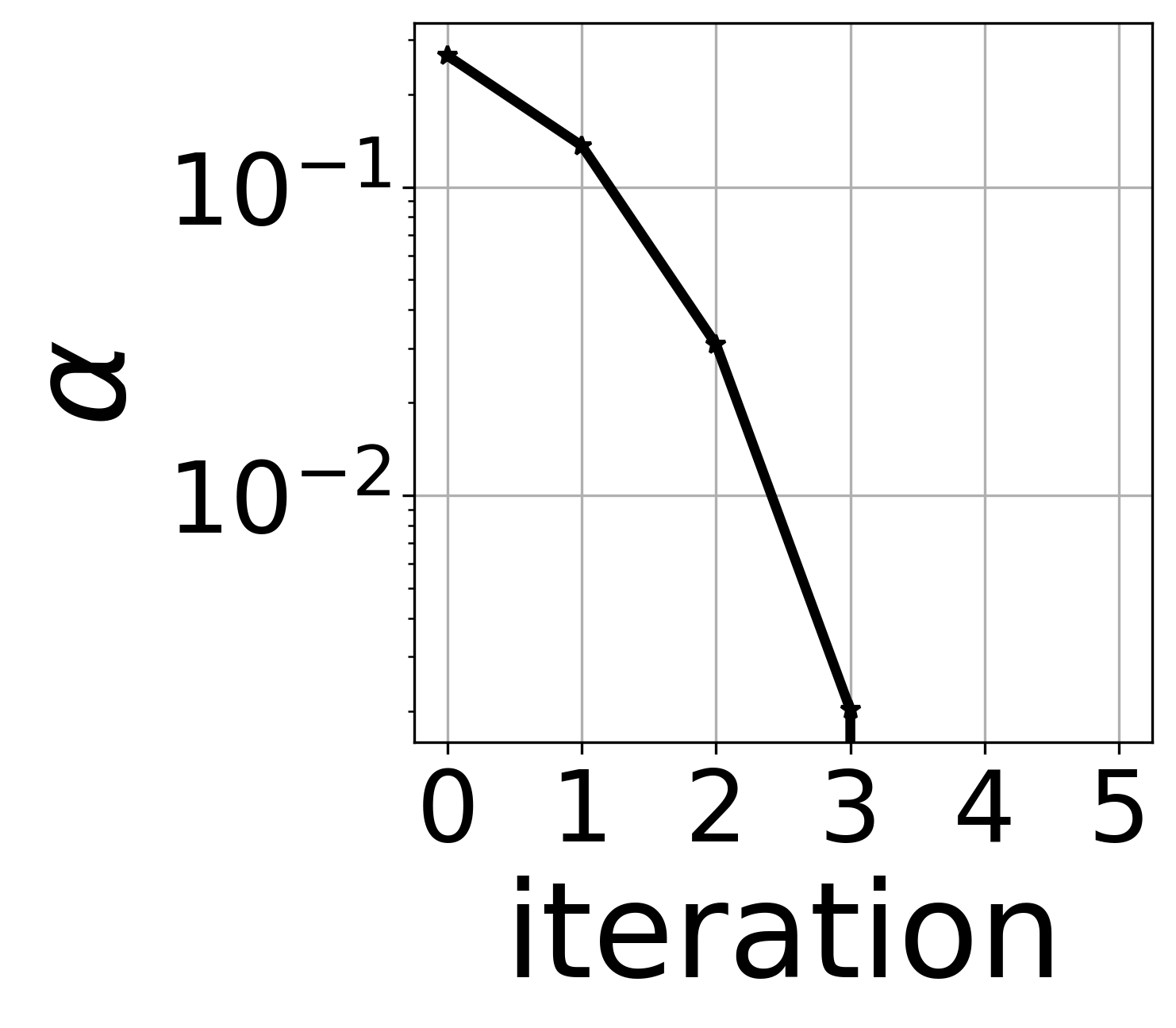}
         \caption{}
     \end{subfigure}          
     \caption{(a) Definition of operator $J$. (b) MC Definition of operator $R$ for $P(Y_1|X_1)=.95$. (c) Definition of fixed-point process \ref{eq:iter_alpha} from (a) and (b). (d) shows convergence in simulation for an initial point $\alpha^{k=0}=.27$ determined by $J$ and initial conditions $\theta$.}
     \label{img:J_R}
\end{figure}

In this work, we derive the difference in mutual information, $\delta^a$, in the augmented dataset using the augmented conditional probabilities in equations \ref{eq:new_augmented_cond}. Notice these conditional probabilities are dependent on the initial conditional and marginal probabilities of $X_1$, $X_2$, and $Y_1$ in the training dataset. We define these as $\theta$. The conditional probabilities in the augmented dataset are also dependent on our helpful error $\beta$, and the error rate of the selector $\alpha^{k}$. Going forward, we will assume $\beta = 1-\alpha^{k}$ because of the counterfactual generation process in Section Implementation where we randomly sample from a set of candidate rationales which were produced with an error $\alpha^{k}$. The  properties of our augmented dataset are defined by our counterfactual generation process, equations \ref{eq:new_augmented_cond}, and the initial conditions in the training dataset. We will call the transform from the initial dataset to the augmented dataset operator $J$. $J$ is shown in Figure \ref{img:J_R} for our initial conditions and varying $\alpha^k$.

\begin{equation}\label{eq:J}
    \delta^a = J(\alpha^{k},\theta)
\end{equation}

Now, from Equation \ref{eq:R}, we have:
\begin{equation}\label{eq:alpha_a}
    \alpha^{k+1} = R(\delta^a) = R(J(\alpha^{k},\theta))
\end{equation}

Now, with our definitions of operators $R$ and $J$, we can show convergence of Algorithm \ref{alg:iter_cda}.

\begin{theorem} \label{thm:convergence}
Process \ref{eq:iter_alpha} converges to expected error $\alpha^{k+1} = R(\delta^a=I(X_1,Y_1),n)$ under Algorithm \ref{alg:iter_cda} for rationale scheme \ref{def:binary_rat}, an $n$ such that $R^{-1} <  J$ for some $\alpha \in [R(\delta^a=I(X_1,Y_1),n),\alpha_T)$, and an initial $\alpha^{k=0}$ such that $R(\delta^a=I(X_1,Y_1),n) <= \alpha^{k+1} < \alpha_T$.
\end{theorem}

Our first condition on the rationale selector is driven by a sufficiently large $n$ such that there is some region of $\alpha^k$ such that $\psi(\alpha^k)$ lies below the line $\alpha^k = \alpha^{k+1}$. This is the region where $\alpha$ iterations are decreasing.
Our second condition on the error rate of the initial selector $\alpha^{k=0}$ ensures that we start in this decreasing region. We also know that $\psi(\alpha^k)$ is monotonic because it is a composition of two monotonic functions.

Finally, notice that our final error is lower bounded by $R(\delta=I(X_1|Y_1),n)$ which is the expected error when there is no mutual information $I(X_2|Y_1)$. This is how well the rationale selector is expected to perform when there is no spurious mutual information in the dataset. With these conditions, we can see that our iteration \ref{eq:iter_alpha} is monotonically decreasing and lower bounded and therefore converges to $\alpha^{k+1} = R(I(X_1|Y_1),n)$ \cite{Bibby_1974}.

Theorem \ref{thm:convergence} guarantees convergence for an initial $\alpha$ in the region such that $\psi(\alpha^k)$ is less than the line $\alpha^{k}=\alpha^{k+1}$. The $\alpha^k$ where $\psi(\alpha^{k})$ crosses that line is our $\alpha_T$. Above $\alpha_T$, our iteration is actually increasing toward another fixed point at $R(\delta^a(-I(X_2,Y_1),n)$ which is the augmented dataset where there is no mutual information $I(X_1,Y_1)$.

To demonstrate an example in the binary setting, we will assume $p(Y_1|X_1)=.9$, $p(Y_1|X_2)=.85$, $p(X_1)=p(X_2)=p(Y_1)=\frac{1}{2}$, and our selector has 35 observations. Notice $X_1$ is our MMI solution. Figure \ref{img:J_R} shows example relations for operator $J$, $R$, the map from $\alpha_{k}$ to $\alpha_{k+1}$, and finally the $\alpha$ convergence through iterations. Note, for these initial conditions, algorithm \ref{alg:iter_cda} converges \textbf{quadratically}. Relation $R$ is computed through simulation. See Appendix Additional Notes on ICDA Convergence 
for more information on Theorem \ref{thm:convergence} and examples under more initial conditions. 

\section{Implementation} \label{sec:implementation}
Algorithm \ref{alg:iter_cda} outlines our iterative approach to counterfactual data augmentation. In this section, we will detail our implementation and make ICDA work in practice. First, we are iterating on the MMI implementation of the rationale scheme, so the first or zeroth iteration of ICDA is vanilla MMI rationalization. Subsequent iterations are the MMI scheme but on different datasets, specifically new counterfactually augmented datasets with each iteration. 

In this work, we focus on rationalization over sentences. This scheme was actually used in the seminal work \cite{chen_learning_2018}, but most other works have focused on rationalization over tokens \cite{lei_rationalizing_2016} \cite{yu_rethinking_2019} \cite{chang_game_2019}. We use the hierarchical transformer network \cite{pappagari_hierarchical_2019} shown in Figure 3 in Appendix Implementation
. A token level transformer encodes each sentence into a representation, and a sentence level transformer operates over the sentence representation. We apply the rationale criteria over the sentence representation. Sparsity is a desiderata of rationales \cite{lei_rationalizing_2016}, so one can choose any of the sparsity constraints at this point. Our experiments use one sentence as the rationale, but this scheme does generalize to rationales over multiple sentences. After the rationale sentence is selected, the sentence representation produced by the token level transformer is left unmasked for the selected rationale, and we mask vectors of sentences not selected in the rationale. We then apply a pooling operation, max-out in our case, and perform classification using the selected sentence rationale representations.

\citeauthor{plyler_making_2021} generated counterfactual documents following a process that depended on the context in unmodified portion of the document. 
Their method relied on training generative models that worked with the initial rationale selector. To generalize to our iterative scheme, we could naively train new generative models for each CDA iteration that worked with the the rationale selector at each iteration, but this would obviously be computationally expensive. Alternatively, we follow a methodology that is more closely related to \cite{zeng_counterfactual_2020} where we simply shuffle rationales between documents to generate the counterfactual samples. This means our counterfactual generation process no longer explicitly respects the context of the rest of the document, $X_2$.

Specifically, for a given rationale selector $S$, and dataset $D$, we generate a set of rationales $A$. We divide this set by the class label of the source documents, so for a binary classification problem, we have $A_0$ and $A_1$. To generate a counterfactual, we sample from the set with the flipped source label. Our generation process becomes:
\begin{equation}
\label{eq:new_cdaproc}
Y_1^c \leftarrow 1-Y_1; \; \; \; \; X_1^c \leftarrow A|(1-Y_1)
\end{equation}
We sample new counterfactual documents during each epoch of training on the augmented datasets. We should also point out that this sampling approach is appropriate and works because we are rationalizing at the sentence level. In the beer review context, swapping one smell sentence for another smell sentence tends to produce coherent documents. If we were to rationalize at the token level, naively swapping rationales would most likely produce incoherent samples, and one would need to consider a generative approach like that used by \cite{plyler_making_2021} or one using more modern LLMs \cite{li_prompting_2024}.

We limit our augmented dataset size to be of equal size to the original dataset. We include original samples in the augmented dataset where the classifier had the lowest prediction error. Motivated by the analysis on helpful $\beta$ errors, we know during counterfactual generation, it is helpful to insert a rationale of the correct aspect even if the rationale on the original document was incorrect. We therefore limit our rationale set for counterfactual generation $A$ to be sourced from only documents where we made a correct prediction, and we take the 10\% rationales, per class,  where the model was most confident in correct prediction. We found that this limited rationale set helped the dev set classification loss converge in some cases.

During the iteration in Algorithm \ref{alg:iter_cda}, $train\_selector$ is a key operation that produces a rationale selector $S$. In our implementation, we train three rationale selectors by varying the random seed across the three runs. During the zeroth iteration, we select the rationale model with the best loss on a dev set. For future iterations, on counterfactual data, we do the three runs over the random seeds and another run that is initialized by the rationale selector from the previous run and fine-tuned on the counterfactually augmented dataset. For the first counterfactual iteration, we select the model with the lowest loss on the augmented dev dataset. We want the iterative process to converge to a stable set of rationales. To encourage this, we select the model with the lowest loss that also has a decreasing rate of change in the rationale set from the previous iteration. If $A^k$ defines the rationales produced during iteration $k$, the change in the rationale set is defined as $\Delta A = \frac{|A^{k+1}-A^K|}{|A^{k+1}|}$. After the first counterfactual iteration. We are seeking models that have a decreasing change in the rationale set: $\Delta A^{k+1} < \Delta A^k$. We define convergence of the iterative procedure, or the stopping criteria, as a selection of the rationale model initialized by the model from the previous iteration that is not improved by training on the augmented dataset. A more detailed version of our iterative algorithm is shown in Appendix Details.

\section{Experiments}
In this work, we adopt two common benchmark datasets for the rationale problem: RateBeer \cite{mcauley_learning_2012} and Tripadvisor \cite{wang_latent_nodate}. These are two multi-aspect datasets where reviews describe beers and hotels respectively, but most reviews cover multiple aspects of the product. 

For the RateBeer dataset, reviews typically describe the appearance, smell, taste, palatability, and overall rating of the beer. \citeauthor{mcauley_learning_2012} annotated about 1,000 reviews assigning each sentence in the review to an aspect label(s). These annotations have been used to evaluate the rationale alignment in \cite{lei_rationalizing_2016} and many works thereafter. The annotations are strictly in the test split of the data and are not used for hyper-parameter tuning. The key metric in rationale works, and this work, is the precision of the machine generated rationale relative to the ground-truth human annotated rationales. This measures the rate the machine generated rationales are also selected by human annotators.
We follow the common practice of evaluating our methods on the appearance, smell, and palatability aspects \cite{lei_rationalizing_2016} \cite{chang_game_2019}. 

The TripAdvisor dataset is much like the RateBeer dataset, but instead of smell, taste, etc., the reviews describe aspects like the location, service, and cleanliness of the hotel. The dataset was originally curated by \cite{wang_latent_nodate} and human-labeled rationales were collected by \cite{bao_deriving_2018}. Again, these rationales are in the test set only. Note there is a separate dataset for each aspect in the beer and hotel datasets.

We also evaluated ICDA on two LLM-generated datasets. Note, a popular paradigm in machine learning now is to generate a synthetic dataset using an LLM and training a classifier on the generated dataset. This saves an engineer the effort of curating a dataset traditionally, and it allows the engineer to deploy a much small, cheaper classifier as compared the LLM. We show that ICDA can fit into this paradigm with two example tasks: blue-tooth headphone reviews "Connectivity" and restaurant reviews "Restaurant". For the "Connectivity" dataset, we prompt the LLM in such a way that the connectivity aspect of the headphones are most informative of the label. For the "Restaurant" dataset, we did not prescribe a desired rationale, but we found that most models converged to a "food" aspect. Exact details for generating the datasets, and statistics relating to all datasets, are in Appendix Data.

We compare our iterative approach to three baselines from the literature. Note, these baselines have been re-implemented so that they align with our studied rationale criteria for fair comparison. The maximum mutual information (MMI) method from \cite{lei_rationalizing_2016} \cite{chen_learning_2018} is our first key baseline. This can be considered the base approach for the CDA and ICDA approaches. We also implement the complement control method from \cite{yu_rethinking_2019}. Note that this re-implementation is the same as MMI+minimax and we did not use the introspective model. We are iterating on a version of counterfactual data augmentation from \cite{plyler_making_2021}, so we of course include one CDA iteration as a baseline. Specifically, during our ICDA runs, we take the iteration-0 runs as the MMI implementation, the iteration-1 runs as the CDA implementation, and the last iteration as the ICDA run. This keeps the implementation details consistent between the methods and we can fully see the affect of the iterations. 

\section{Results}
\begin{table*}[t]
\centering
\begin{tabular}{lccc|ccc}
\hline
& Appearance & Smell & Palatability & Location & Service & Cleanliness \\
\hline
 COMP & 56.4 \tiny{$\pm$ 11.3 } & 50.8 \tiny{$\pm$ 4.0 } & 30.9 \tiny{$\pm$ 5.9} & 60.3 \tiny{$\pm$ 6.9 } & 64.7 \tiny{$\pm$ 4.0 } & 50.5 \tiny{$\pm$ 7.2} \\
 MMI & 47.5 \tiny{$\pm$ 21.3 } & 56.0 \tiny{$\pm$ 10.8 } & 23.1 \tiny{$\pm$ 3.9 } & 71.0 \tiny{$\pm$ 4.1 } & 66.8 \tiny{$\pm$ 4.3 } & 55.8 \tiny{$\pm$ 2.7} \\
 CDA & 62.9 \tiny{$\pm$ 23.1} & 77.8 \tiny{$\pm$ 11.9} & 24.7 \tiny{$\pm$ 5.4} & 84.7 \tiny{$\pm$ 2.2} & 70.0 \tiny{$\pm$ 5.4} & 53.9 \tiny{$\pm$ 15.7} \\
 ICDA & \textbf{66.6} \tiny{$\pm$ 17.5} & \textbf{93.6} \tiny{$\pm$ 2.8} & \textbf{37.6} \tiny{$\pm$ 21.0} & \textbf{88.0} \tiny{$\pm$ .8} & \textbf{74.0} \tiny{$\pm$ 4.5} & \textbf{56.0} \tiny{$\pm$ 17.1} \\
\hline
\end{tabular}
\caption{\label{tbl:real_results}
Rationale Precision on the test set on real data. The appearance, smell, and palatability datasets are sourced from the beer and hotel datasets. Mean and standard deviation are reported.
}
\end{table*}
Table \ref{tbl:synth_results} shows the results on the LLM generated datasets. Our method, ICDA, showed an improvement over all baselines. The ICDA results here are further iterations on the MMI and CDA results. Based on the theory in Section Iterative Counterfactual Data Augmentation, we should expect CDA to show an improvement over MMI and ICDA to show another improvement over CDA. This was true for the Connectivity dataset, but on average, it wasn't true for the Restaurant dataset. The convergence plots are shown in Appendix Additional Results 
and show that our experiments converged to the "Food" aspect on the restaurant dataset $\frac{2}{3}$ times and the drop in average rationale precision for CDA is a product of that $\frac{1}{3}$ experiment converging to another aspect. Remember, for this restaurant dataset, we did not necessary prescribe a "desired" aspect apriori.

Table \ref{tbl:real_results} shows the rationale precision results on the human generated and annotated test datasets. Again, our ICDA method outperformed the baselines on datasets when averaged over the three runs of the experiment. To see how each run converged, see the charts in Appendix Convergence Plots.

\begin{table}
\centering
\begin{tabular}{lcc}
\hline
& Connectivity & Restaurant \\
\hline
COMP & 56.9 \tiny{$\pm$ 4.9 } & 41.3 \tiny{$\pm$ 7.5 }  \\
 MMI & 50.3  \tiny{$\pm$ 8.9 } & 54.6 \tiny{$\pm$ 5.0}  \\
 CDA & 57.9 \tiny{$\pm$ 7.6} & 51.5 \tiny{$\pm$ 11.7}  \\
 ICDA & \textbf{63.2} \tiny{$\pm$ 5.2} & \textbf{57.4} \tiny{$\pm$ 17.2 }  \\
 
\hline
\end{tabular}
\caption{\label{tbl:synth_results}
Rationale Precision on the test set on LLM generated data. Mean and standard deviation over are reported.
}
\end{table}

\section{Related Work}
The concept of rationales was created in \cite{lei_rationalizing_2016} and its relation to mutual information was shown by \cite{chen_learning_2018}. There have been a variety of follow-up works tackling different issues faced by these networks. \citeauthor{yu_rethinking_2019} addressed the problem of rationale degeneration through the idea of complement control: minimizing the amount of information left in the complement of the rationale. \citeauthor{chang_game_2019} leveraged the idea of counterfactuals for rationales, but did these counterfactuals were different selections over a single input document. \citeauthor{chang_invariant_2020} used a variety of training dataset environments and invariant learning to find a rationale policy that generalizes across the domains. Our work builds most directly on \citeauthor{plyler_making_2021} which uses generative models to create a counterfactual dataset. \citeauthor{liu_d-separation_2023} and \citeauthor{zhang_towards_2023} have both taken a causal prospective on the rationalization problem where \cite{liu_d-separation_2023} builds counterfactuals during training by perturbing the rationales and \citeauthor{zhang_towards_2023} evaluated differences in predictions using the whole document versus the rationales. We selected MMI \cite{lei_rationalizing_2016} \cite{chen_learning_2018} and CDA \cite{plyler_making_2021} as methods to re-implement as baselines because our work is most directly built on these methods. We also re-implemented \cite{yu_rethinking_2019} because it directly generalizes to the rationalization over sentences setting while many methods are tied to the token-rationalization scheme. Its also important to point out that our data augmentation method is offline from model training and the $train\_rationale$ procedure in Algorithm \ref{alg:iter_cda} can generally be replaced with any rationalization strategy.

Prior work in the causal modeling community focused on identifying causal signals and helping models ignore spurious signals. Sun et al. 2021 shows that identifying such causal signals helps models become shift-invariant. Veitch et al. 2021 leverages causal inference and  proposes the idea of counterfactual invariance as a model requirement and training strategy for avoiding spurious correlations.

Counterfactual data augmentation was introduced as a method for controlling gender bias in datasets \cite{lu_gender_2020}. They hand-crafted a strategy for changing gender carrying terms in the datasets in such a way that the augmented dataset would have less gender bias. This idea has been built upon in the literature. \citeauthor{zeng_counterfactual_2020} created counterfactuals by swapping named entities in a training dataset, and we should note our method of shuffling rationales between documents is similar counterfactual generation process. \citeauthor{kaushik_learning_2020} used human annotators to generate counterfactual datasets and showed that this can help downstream models generalize out-of-domain and \citeauthor{deng_counterfactual_2023} adapted this strategy to active learning. \citeauthor{li_prompting_2024} explored prompting LLMs for counterfactual generation.

\section{Impact, Limitations, and Conclusions} 
While the datasets studied were relatively benign, one can imagine this as a methodology for controlling more serious spurious signals in training datasets. To that end, we should also acknowledge that this method relies on generating counterfactuals without human intervention or annotation. For errant cases, ICDA could produce counterfactual documents that are not factual, and a user should be aware of the dataset they are using and how the algorithm could manipulate that data.

Our theory in the binary section assumed the rationale model at iteration $k$ will have the expected error from $R$ and is not inherently random. The NLP rationale learning problem is inherently stochastic. \citeauthor{lei_rationalizing_2016} framed this a reinforcement learning problem where the selector is our agent and the hard selection over tokens are the available actions. It is often the case that our agent converges to a point where the rationales are low quality. We therefore selected the best MMI model, according to dev set prediction loss, from three runs with different random seeds as a our initial rationale selector for future iterations. 
Tables \ref{tbl:real_results} and \ref{tbl:synth_results} show the MMI results vary. This noise is not factored into the theoretical framing for the binary case. 

The ICDA approach is more computationally expensive than the baselines because it is running the baselines multiple times. Figure \ref{img:J_R} suggests ICDA should converge quadratically. Across all experiments, Figure 4 in the Appendix shows ICDA converged in five or less iterations. We always trained a model to convergence on each augmented dataset. It might be possible to save computation, and allow ICDA to see better counterfactuals earlier, by not training to convergence during the earlier iterations.

This work presents an iterative approach to counterfactual data augmentation. We show how a process that starts with noisy interventions can self correct and converge to a process with less noise. 
Specifically, we showed how initial rationale models that aligned relatively poorly with human annotations could be iteratively improved through a CDA scheme that does not rely on human annotation, domain specific knowledge, or generative models. Our iterative approach outperformed the baselines for both human sourced datasets as well as LLM generated datasets. 

\section{Acknowledgments}
This research was supported by the NSF Grants: Integrated Data-driven Technologies for Individualized Instruction in STEM Learning Environments (1726550), CAREER: Improving Adaptive Decision Making in Interactive Learning Environments (1651909), and Generalizing Data-Driven Technologies to Improve Individualized STEM Instruction by Intelligent Tutors (2013502).

\bibliography{itercda.bib}

\appendix \label{app:theory}

\appendix \label{app:implementation}

\section{Additional Notes on ICDA Convergence}
We presented a rationale scheme \ref{def:binary_rat} where the expected error is dependent on the difference in mutual information. To see this, consider the case where $X_1$ and $X_2$ are equally informative of the label, $I(X_1,Y_1)=I(X_2,Y_1)$. Both $X_1$ and $X_2$ are equally likely to satisfy the MMI condition, relation \ref{eq:mmi}, and if the rationale constraints meant the model selected either $X_1$ or $X_2$, we would expect $X_1$ and $X_2$ to be selected at equal rates and our error rate would be $\alpha=\frac{1}{2}$. Now consider the case where $X_2$ is not informative of $Y_1$. In this case, we would expect the rationale selector to make very few mistakes and it should reach toward the optimal solution $X_M \leftarrow X_1$ and $\alpha=0$. As we change from the equally informative case to the non-informative case, we are increasing $\delta$ and $\alpha^{k+1}$ is increasing. Therefore, we can see $\alpha^{k+1}$ and $\delta$ to be inversely related though operator $R$.

In Section Iterative Counterfactual Data Augmentation, $X_1$, $X_2$, and $Y_1$ are approximated by binary variables. Remember, the rationalization model \cite{lei_rationalizing_2016} is a Monte Carlo estimate of the maximum mutual information criteria \ref{eq:mmi} which is effectively evaluating $I(Y_1|X_1)$ and $I(Y_1|X_2)$ and choosing the maximum. In the binary setting, identifying whether $X_1$ or $X_2$ is the maximizer is algorithmically trivial given sufficient observations. We therefore use rationale scheme \ref{def:binary_rat} in this analysis. To make the analogy closer to our NLP rationalization problem, consider the scenario where we have a limited number of observations, $n$, of variables $Y_1$, $X_1$, and $X_2$ and our rationale model is tasked with selecting the signal, either $X_1$ or $X_2$ that satisfies the maximum mutual information criteria. This would be the variable that has the maximum agreement with $Y_1$. In the limited observation case, for example $n=35$, we define operator $R$ through 60,000 simulations in Figure \ref{img:J_R}. We break ties by randomly selecting either $X_1$ or $X_2$. This is our expected error for the next round of CDA $\alpha^{k+1}$ for a given difference in mutual information $\delta$. In this case, our sequence of $\alpha$ converges to 0 if the initial $alpha$ is in the region such that $\delta^a$ is positive. Interestingly, $\delta^a=0$ is also a fixed point but is not an attractor and corresponds to $\alpha=\frac{1}{2}$ if $X_1$ and $X_2$ are initially equally informative of $Y_1$. For the case in Figure \ref{img:J_R}, the fixed point is not $\frac{1}{2}$ and is determined by the initial probabilities in the dataset and $n$. 
Alternatively, we could have an $n$ that does not satisfy our requirement $R^{-1} < J$. In this case, our sequence of $\alpha$ converges to $\alpha=\frac{1}{2}$ when $X_1$ and $X_2$ are initially equally informative of $Y_1$.

\section{Handling Large Datasets}
In the context of larger datasets, notice ICDA’s run time is dominated by training a new rationale selector at each iteration. Generating the counterfactuals is very cheap. For very large datasets, it might be possible to sample a subset of the dataset for training the rationale selector. As long as that subset is sufficient for getting a quality estimate of the most informative signal, you could still do the cheap part, debiasing the entire dataset through counterfactual generation, on the entire dataset while benefiting from training on the smaller subset.

\section{Detailed Iterative CDA Implementation} \label{sec:detailed_icda}
Algorithm \ref{sec:detailed_icda} shows a more detailed version of our iterative procedure.

\begin{algorithm}[h!]
\caption{Iterative CDA Procedure}\label{alg:iter_cda}
\begin{algorithmic}
\Require $D$ is a dataset with documents $X$ and labels $Y$. $rand\_seeds$ is a list of integers.
\State $D' \gets D$
\State $rand\_seeds' \gets rand\_seeds$
\State $d \gets 1$
\State $k \gets 0$
\While {not converged}
    \State $L \gets 1e10$
    \For{$r$,$rand\_seed$ in enumerate($rand\_seeds'$)}
      \If {$r > len(rand\_seeds)$}
         $S_0 \gets S$
      \Else{ }
        $S_0 \gets random\_init$        
      \EndIf  
      \State $S_i,L_i,A_0,A_1 \gets train\_selector(D',S_0)$
      \If {$k>1$ and  $(set\_change(A_0,A_0')+set\_change(A_1,A_1'))/2 < d$}
         $M_i \gets 1$
     \Else{ } 
       $M_i \gets 1e10$
     \EndIf
      \If{$L_iM_i < L$}
         \State $S \gets S_i$
         \State $L_i \gets L_i M_i$
         \State $best\_seed \gets rand\_seed$
      \EndIf
    \EndFor
    \State $D^c \gets infer\_counterfactuals(D,S)$ 
    \State $D^a \gets concatenate(D,D^c)$
    \State $D' \gets D^a$
    \State $rand\_seeds' \gets rand\_seeds.append(best\_seed)$
    \State $k \gets k+1$
    \If{$r > len(rand\_seed$ and $best\_epoch=-1$} 
      beak
     \EndIf
     \Comment{$best\_epoch=-1$ means the model did not improve classification loss }
\EndWhile
\end{algorithmic}
\end{algorithm}

\section{Network}

\begin{figure*}[h]
\includegraphics[width=0.8\textwidth]{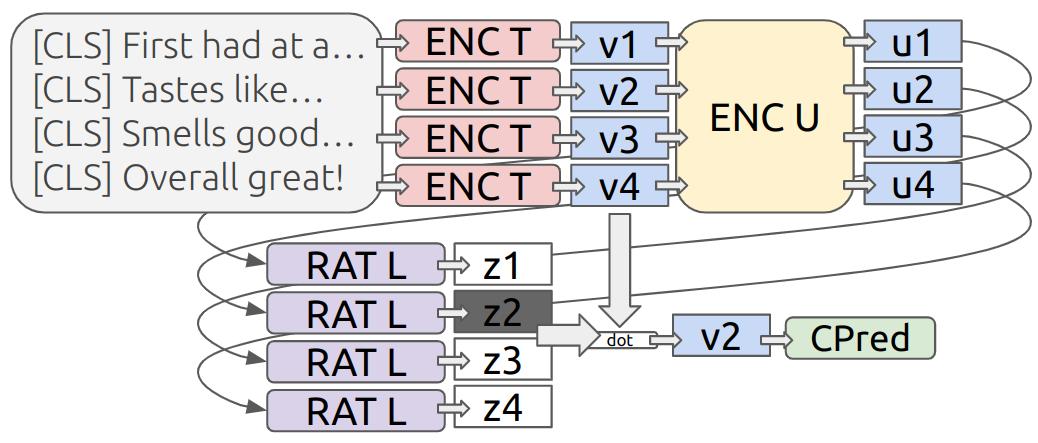}
\centering
\caption{Network configuration for rationalization over sentences with a hierarchical transformer.}
\label{img:network}
\end{figure*}

Figure \ref{img:network} shows the network configuration for all models and baselines. A BERT model encodes each sentence, another encoder takes the sentence vectors as input, we apply hard rationalization over the sentences, and a prediction is made using the sentence rationale.

\section{Data Details} \label{sec:data_details}
For the Beer data, we use the "correlated" splits from \cite{plyler_making_2021}. This is supposed to be the beer data as it would appear in the 'wild' and the aspects are not de-correlated through their label. The ratings are binarized by treating ratings $>=.6$ as positive and ratings $<=.4$ as negative. We use the hotel splits and annotations from \cite{bao_deriving_2018}. We down sample the size of the clean and service aspects to be about the same size as the location split to save computational resources.

We modify all training and dev datasets by deduplifying, filtering all samples with less than five sentences, and balancing the classes. We use spacy \cite{spacy2} to segment the sentences for all datasets except the beer datasets. For the beer datasets, the rationale annotations were originally at the sentence level, we therefore tried to create a sentence segmenter that matched the segmenter used during those annotations. 

The beer annotations were originally at the sentence level, so we were generally able to use those annotations directly in this work. For the hotel datasets, we treat a sentence as "ground-truth-rationale" if there is a word in that sentence that was annotated as such. For the "connect" dataset, we treat a sentence as "ground-truth-rationale" if any of the strings "connect", "cuts out", or "dropout" appear in the sentence. For the "restaurant" dataset, we treat a sentence as "ground-truth-rationale" if one of the strings "food", "dish", "flavor", "delicious", "bland", "cook", or "ingredients" appear in the sentence.

\subsection{Generating the "Connectivity" Dataset}
We used meta-llama:Llama-2-7b-chat-hf \cite{touvron_llama_2023} to generate this dataset. We used the prompt: "'please a write review between 8 and 15 sentences for a pair of bluetooth headphones. mention the look, battery, weight, sound, microphone, sturdyness, comfort, \$ADJECTIVE connectivity. start the review with: "My review:"". Where \$ADJECTIVE is from the list: 'bad', 'not good','ok', 'pretty good','very good', 'excellent', 'acceptable', 'satisfactory', 'reliable', 'unreliable', 'spotty', 'terrible'. We map each of these adjectives to a positive or negative sentiment. We used some keyword matching to remove sentences that contained part of the prompt.

\subsection{Generating the "Restaurant" Dataset}
We followed the same settings as the "connectivity" dataset with a different prompt for generating the "restaurant" dataset. Prompt: 'please write a detailed review with at least five sentences for a new restaurant. your review should be a \$RATING/10 but do not explicitly state that in the review. Start the review with the phrase "My Review:"'. Where \$Rating is a number between 1 and 10. We tried to remove sentences that talked about the overall rating of the restaurant so we filtered sentences with the one of the strings "review", "give it a", or "out of".

\begin{table*}
\centering
\begin{tabular}{lccccccccc}
\hline
& \multicolumn{2}{c}{Train} & \multicolumn{2}{c}{Dev} & \multicolumn{2}{c}{Test} \\
\hline
Aspect & Doc. Co. & Sent. Co. & Doc. Count & Sent. Co. & Doc. Co. & Sent. Co. \\
\hline
Appearance & 29642 & 10.95 \small{$\pm$ 5.30} & 3478 & 10.23 \small{$\pm$ 4.63} & 955 & 9.32 \small{$\pm$ 2.83} \\
Smell & 26874 & 10.92 \small{$\pm$ 5.31} & 3534 & 10.24 \small{$\pm$ 4.56} & 917 & 9.38 \small{$\pm$ 2.82} \\
Palate & 24174 & 10.99 \small{$\pm$ 5.38} & 3528 & 10.43 \small{$\pm$ 4.83} & 823 & 9.51 \small{$\pm$ 2.83} \\
\hline
Location & 10144 & 9.55 \small{$\pm$ 3.60} & 1470 & 9.78 \small{$\pm$ 3.70} & 200 & 8.76 \small{$\pm$ 3.93} \\
Service & 10000 & 9.68 \small{$\pm$ 3.76} & 1000 & 9.47 \small{$\pm$ 3.53} & 199 & 8.44 \small{$\pm$ 4.00} \\
Clean & 10000 & 9.45 \small{$\pm$ 3.61} & 1000 & 9.55 \small{$\pm$ 3.81} & 196 & 8.27 \small{$\pm$ 4.07} \\
\hline
Connectivity & 7500 & 9.97 \small{$\pm$ 4.02} & 1200 & 9.99 \small{$\pm$ 3.90} & 1200 & 10.03 \small{$\pm$ 4.00} \\
Restaurant & 4574 & 6.31 \small{$\pm$ 1.93} & 626 & 6.26 \small{$\pm$ 1.81} & 582 & 6.28 \small{$\pm$ 1.88} \\
\hline
\end{tabular}
\caption{\label{tbl:datasets}
Dataset Statistics
}
\end{table*}
\section{Settings}
\subsection{Rationalization Settings}
The rationale networks of course make a hard selection over units in the input. The non-differentiability of this hard selection is a core challenge training these networks. Across all our evaluated methods, we implemented the straight-through method for getting past this non-differentiable \cite{bengio_estimating_2013}. Other rationale works have used Gumbel-softmax\cite{jang_categorical_2017} or the REINFORCE algorithm \cite{williams_simple_2004}.
The hyperparameters of the rationale networks were generally the same across the considered methods. All models were training with a AdamW \cite{loshchilov_decoupled_2019} optimizer with a learning rate of 1e-6, a batch size of 64, a weight decay of 1e-2, a token level encoder initialized from BERT with six layers \cite{devlin_bert_2019}, and randomly initialized sentence level encoder initialized with four transformer layers. The COMP model \cite{yu_rethinking_2019} has an additional $\lambda_{comp}$ parameter which was chosen from a set based on the dev set loss. These parameters were generally chosen to improve the dev set class prediction loss converge for the MMI model.

To generate the counterfactual datasets, we accept rationales into the counterfactual sets that are in the lowest 10\% of prediction error. We also only take original documents into the augmented dataset were the model made a correct prediction. These parameters were chosen based on a small experiment on palate dataset after observing the rationale model would not learn on a counterfactual dataset with all source samples. Choosing 10\% helped the prediction loss converge on the augmented dev set. We also limit the size of the augmented dataset to match the size of the original dataset to save computational resources. We take source samples with the smallest prediction error.

Across all methods and datasets, we completed three runs of the experiment and report the mean and standard deviations across these runs. For each run, and for the MMI, CDA, and ICDA methods, we select the model with the best loss across three random seeds. For the CDA and ICDA methods, we also include a model that is initialized from the previous iteration as a candidate for selection. For the COMP model, we select the best model based on dev set classification loss across a grid search with two random seeds and the $\lambda_{comp}$ set of [.5,1].

\subsection{Avoiding Rationale Degeneration}
A well studied problem for rationale networks is degeneration \cite{yu_rethinking_2019}. This is the scenario where the two components of the rationale network collude and the rationale selector passes a signal to the classifier for the class decision. Degenerate models are effectively making a decision over the entire input document instead of just the selected rationale. This violates many of the rationale desiderata \cite{lei_rationalizing_2016}, and most importantly it produces non-sensical rationales that do not align with the desired signal. There are works that seek to prevent degeneration \cite{yu_rethinking_2019}. For the network used in the this work, rationale degeneration was observed in most cases and it tended to be positional. Eventually, the network would select the first sentence for one class, and any other sentence for the other class. To prevent us from choosing a degenerate model or checkpoint, we only take checkpoints where the rationales indices for both classes are roughly the same over all documents in the dev set. We do not take checkpoints were the class-wise rationales diverge on aggregate by more than 20\% in any rationale index bin. We also apply this difference criteria to the first and second rationale index bin because degeneration tended to occur on the first sentence. We developed this methodology, and the convergence criteria, by inspecting how the rationale index bins changed through the iterations on the training and dev sets. We also did some sanity spot checking of rationales and counterfactuals on the training and dev sets. The rationale index bins are created by normalizing the selected rationale index by the total number of sentences in a document and then binning that normalized index into tenth percentile buckets.

\subsection{Convergence Plots} \label{sec:convergence_plots}
\begin{figure}[h!]
\includegraphics[width=0.8\textwidth]{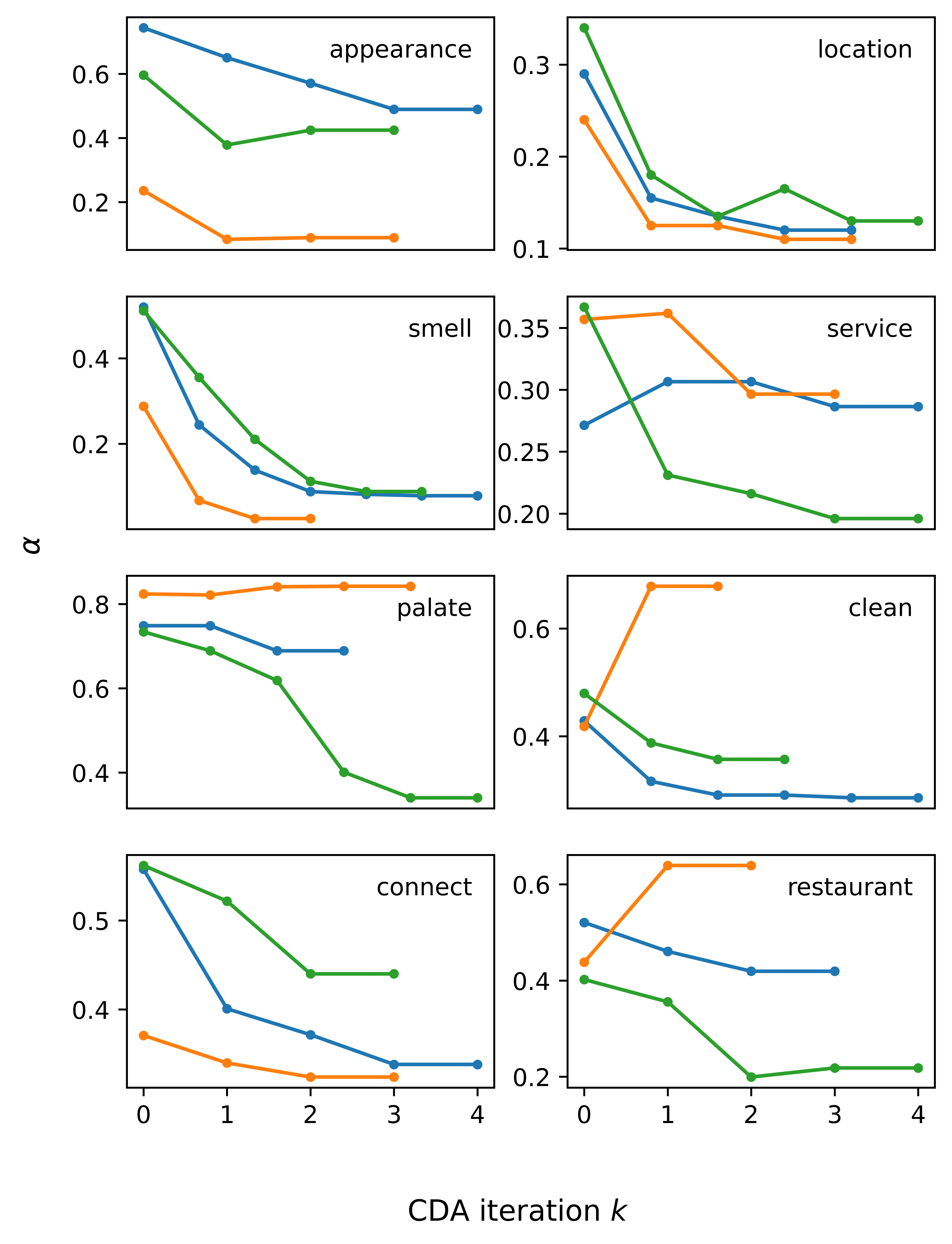}
\centering
\caption{Convergence results over all experiments runs. Note that the general trend for the reationale error to decrease with CDA iterations. Rationale error $\alpha$ is taken as the complement of the rationale precision.}
\label{img:cda_iters}
\end{figure}

Figure \ref{img:cda_iters} shows how the rationale error changed over CDA iterations.

\end{document}